\documentclass[10pt,twocolumn,letterpaper]{article}

\usepackage{cvpr}
\usepackage{times}
\usepackage{epsfig}
\usepackage{graphicx}
\usepackage{amsmath}
\usepackage{amssymb}
\usepackage{upgreek}
\usepackage[utf8x]{inputenc}

\usepackage[pagebackref=true,breaklinks=true,letterpaper=true,colorlinks,bookmarks=false]{hyperref}
\newcommand*\samethanks[1][\value{footnote}]{\footnotemark[#1]}
\cvprfinalcopy %

\def\cvprPaperID{***} %

\ifcvprfinal\fi
\begin{document}

\title{Self-Supervised Learning of 3D Human Pose using Multi-view Geometry}
\author{Muhammed Kocabas \thanks{equal contribution}
\qquad
Salih Karagoz \samethanks[1]
\qquad
Emre Akbas
\\
Department of Computer Engineering, Middle East Technical University \\
{\tt\small \{muhammed.kocabas,e234299,eakbas\}@metu.edu.tr}
}

\maketitle
\begin{abstract}
Training accurate 3D human pose estimators requires large amount of 3D ground-truth data  which is costly to collect. Various weakly or self supervised pose estimation methods have been proposed due to lack of 3D data. Nevertheless, these methods, in addition to 2D ground-truth poses, require either additional supervision in various forms (\eg unpaired 3D ground truth data, a small subset of labels) or the camera parameters in multiview settings. To address these problems, we present EpipolarPose, a self-supervised learning method for 3D human pose estimation, which does not need  any 3D ground-truth data or camera extrinsics. During training, EpipolarPose estimates 2D poses from multi-view images, and then, utilizes epipolar geometry to obtain a 3D pose and camera geometry which are subsequently used to train a 3D pose estimator. We demonstrate the effectiveness of our approach on standard benchmark datasets (\ie Human3.6M and MPI-INF-3DHP) where we set the new state-of-the-art among weakly/self-supervised methods. Furthermore, we propose a new performance measure Pose Structure Score (PSS) which is a scale invariant, structure aware measure to evaluate the structural plausibility of a pose with respect to its ground truth. Code and pretrained models are available at \url{https://github.com/mkocabas/EpipolarPose}

\end{abstract}
\section{Introduction}
\label{sec:intro}
Human pose estimation in the wild is a challenging problem in computer vision. Although there are large-scale datasets \cite{mpii,coco} for two-dimensional (2D) pose estimation, 3D datasets \cite{h36m, monocular3d} are either limited to laboratory settings or limited in size and diversity. Since collecting 3D human pose annotations in the wild is costly and 3D datasets are limited,  researchers have resorted to weakly or self supervised approaches with the goal of obtaining an accurate 3D pose estimator by using minimal amount of additional supervision on top of the existing 2D pose datasets. Various methods have been developed to this end. These methods, in addition to ground-truth 2D poses, require either additional supervision in various forms (such as unpaired 3D ground truth data\cite{tung2017}, a small subset of labels \cite{rhodin2018})  or (extrinsic) camera parameters in multiview settings \cite{pavlakos2017harvesting}. To the best of our knowledge, there is only one method \cite{drover2018} which can produce a 3D pose estimator by using only 2D ground-truth. In this paper, we propose another such method. 
\begin{figure}
\centering
\includegraphics[width=1 \columnwidth]{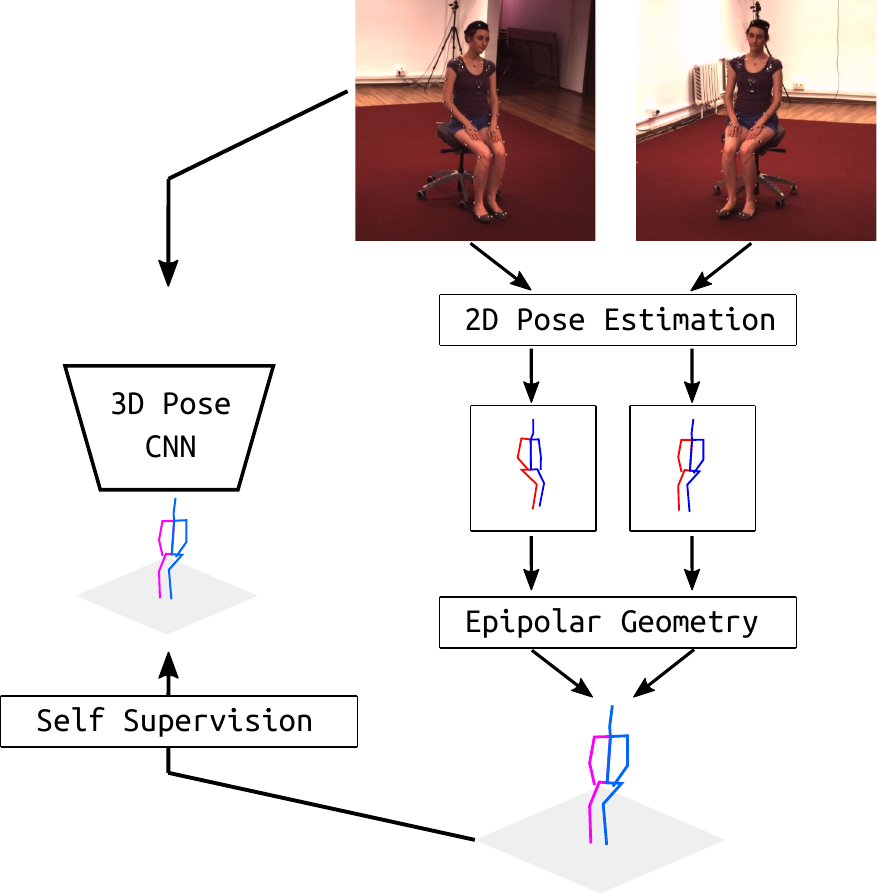}
\caption{\textbf{EpipolarPose} uses 2D pose estimation and epipolar geometry to obtain 3D poses which are subsequently used to train a 3D pose estimator.}
\label{fig:teaser}
\end{figure}

Our method, ``EpiloparPose,’’ uses 2D pose estimation and epipolar geometry to obtain 3D poses, which are subsequently used to train a 3D pose estimator. EpipolarPose works with an arbitrary number of cameras (must be at least 2) and it does not need any 3D supervision or the extrinsic camera parameters, however,  it can utilize them if provided. On the Human3.6M \cite{h36m} and MPI-INF-3DHP \cite{monocular3d} datasets, we set the new state-of-the-art in 3D pose estimation for weakly/self-supervised methods. 

Human pose estimation allows for subsequent higher level reasoning, \eg in autonomous systems (cars, industrial robots) and activity recognition. In such tasks, structural errors in pose might be more important than the localization error measured by the traditional evaluation metrics such as MPJPE (mean per joint position error) and PCK (percentage of correct keypoints). These metrics treat each joint independently, hence, fail to asses the whole pose as a structure. Figure \ref{fig:pose} shows that structurally very different poses yield the same MPJPE with respect to a reference pose. To address this issue, we propose a new performance measure, called the Pose Structure Score (PSS), which is sensitive to structural errors in pose. PSS computes a scale invariant performance score with the capability to score the structural plausibility of a pose with respect to its ground truth. Note that PSS is not a loss function, it is a performance measure that can be used along with MPJPE and PCK to account for structural errors made by a pose estimator. 

To compute PSS, we first need to model the natural distribution of ground-truth poses. To this end, we use an unsupervised clustering method. Let $\mathbf{p}$ be the predicted pose for an image whose ground-truth is $\mathbf{q}$. First, we find which cluster centers are closest to $\mathbf{p}$ and $\mathbf{q}$. If both of them are closest to (\ie assigned to) the same cluster center, then the pose structure score (PSS) of $\mathbf{p}$ is said to be 1, otherwise 0.
\paragraph{Contributions}
Our contributions are as follows: 
\begin{itemize}
\item We present EpipolarPose, a method that can predict 3D human pose from a single-image. For training,  EpipolarPose does not require any 3D supervision nor camera extrinsics. It creates its own 3D supervision by utilizing epipolar geometry and 2D ground-truth poses. 
\item We set the new state-of-the-art among weakly/self-supervised methods for 3D human pose estimation.
\item We present Pose Structure Score (PSS), a new performance measure  for 3D human pose estimation to better capture structural errors.  
\end{itemize}
\section{Related Work}
Our method, EpipolarPose, is a single-view method during inference; and a multi-view, self-supervised method during training. Before discussing such methods in the literature, we first briefly review entirely single-view (during both training and inference) and entirely multi-view methods for completeness. 
\paragraph{Single-view methods} 
In many recent works, convolutional neural networks (CNN) are used  to estimate the coordinates of the 3D joints directly from images \cite{structured3d,fuse_tekin,Tome_2017_CVPR,compositional2017,monocular3d}. Li and Chan \cite{monocular_asia} were the first to show that deep neural networks can achieve a reasonable accuracy in 3D human pose estimation from a single image. They used two deep regression networks and body part detection. Tekin \etal \cite{structured3d} show that combining traditional CNNs for supervised learning with auto-encoders for structure learning can yield good results. Contrary to common regression practice, Pavlakos \etal \cite{pavlakos17volumetric} were the first to consider 3D human pose estimation as a 3D keypoint localization problem in a voxel space. Recently, ``integral pose regression’’ proposed by Sun \etal \cite{Sun_2018_ECCV} combined volumetric heat maps with a soft-argmax activation and obtained state-of-the-art results.

Additionally, there are two-stage approaches which decompose the 3D pose inference task into two independent stages: estimating 2D poses, and lifting them into 3D space \cite{2dposematching,Moreno_cvpr2017,martinez_2017_3dbaseline,posegrammar2018,zhou2016sparseness,2dposematching,Tome_2017_CVPR,monocular3d}. Most recent methods in this category use state-of-the-art 2D pose estimators \cite{cao2017,wei2016,newell2016,kocabas2018} to obtain joint locations in the image plane. Martinez \etal \cite{martinez_2017_3dbaseline} use a simple deep neural network that can estimate 3D pose given the estimated 2D pose computed by a state-of-the-art 2D pose estimator. Pavlakos \etal \cite{pavlakos2018ordinal} proposed the idea of using  ordinal depth relations among  joints to bypass the need for full 3D supervision. 

Methods in this category require either full 3D supervision or extra supervision (\eg ordinal depth) in addition to full 3D supervision. 
\paragraph{Multi-view methods} 
Methods in this category require multi-view input both during testing and training. Early work \cite{amin2013, martin2010,burenius2013, belagiannis2014, belagiannis2016} used 2D pose estimations obtained from calibrated cameras to produce 3D pose by triangulation or pictorial structures model. More recently, many researchers \cite{elhayek2017} used deep neural networks to model multi-view input with full 3D supervision. 
\paragraph{Weakly/self-supervised methods}
Weak and self supervision based methods for human pose estimation have been  explored by many  \cite{drover2018, rhodin2018, tung2017, pavlakos2017harvesting} due to lack of 3D annotations. Pavlakos \etal \cite{pavlakos2017harvesting} use a pictorial structures model to obtain a global pose configuration from the keypoint heatmaps of multi-view images. Nevertheless, their method needs full camera calibration and a keypoint detector producing 2D heatmaps.

Rhodin \etal \cite{rhodin2018} utilize multi-view consistency constraints to supervise a network. They need a small amount of 3D ground-truth data to avoid degenerate solutions where poses collapse to a single location. Thus, lack of in-the-wild 3D ground-truth data  is a limiting factor for this method \cite{rhodin2018}.

Recently introduced deep inverse graphics networks \cite{kulkarni,3dinterpreter} have been applied to the human pose estimation problem \cite{tung2017, drover2018}. Tung \etal \cite{tung2017} train a generative adversarial network which has a 3D pose generator trained with a reconstruction loss between projections of predicted 3D poses and input 2D joints and a discriminator trained to distinguish predicted 3D pose from a set of ground truth 3D poses. Following this work, Drover \etal \cite{drover2018} eliminated the need for 3D ground-truth by modifying the discriminator to recognize plausible 2D projections. 

To the best of our knowledge, EpipolarPose and Drover \etal’s method are the only ones that do not require any 3D supervision or camera extrinsics. While their method does not utilize image features, EpipolarPose makes use of both image features and epipolar geometry and produces much more accurate results (4.3 mm less error than Drover \etal’s method).
\section{Models and Methods}
\begin{figure}
\centering
\includegraphics[width=1\columnwidth]{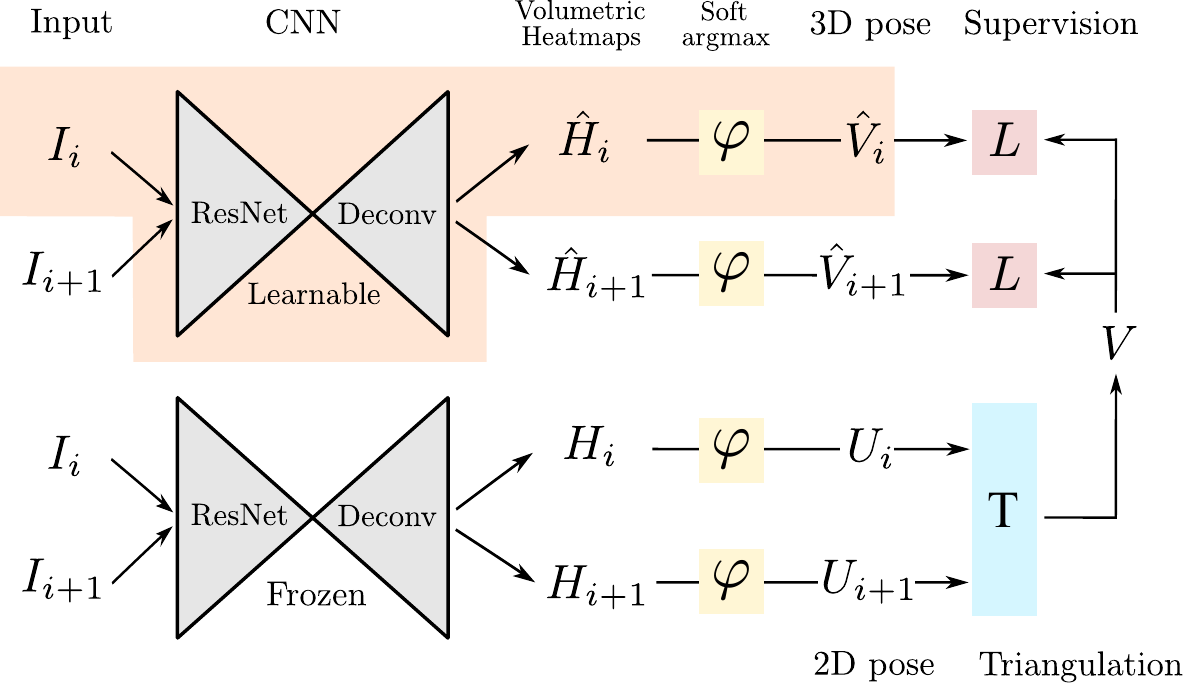}
\caption{\textbf{Overall architecture of EpipolarPose during training.} The orange-background part in the upper branch denotes the inference pipeline. During training, EpipolarPose is multi-view: a pair of images $(I_i, I_{i+1})$ simultaneously taken by two consecutive cameras is fed into the CNN pose estimators. It is also self-supervised: the 3D pose ($V$) generated by the lower branch using triangulation (\ie epipolar geometry) is used as a training signal for the CNN in the upper branch. During inference (the orange-background part), EpipolarPose is a monocular method: it takes a single image ($I_i$) as input and estimates the corresponding 3D pose ($\hat{V}_i$). ($\varphi$: soft argmax function, T: triangulation, L: smooth L1 loss.) }
\label{fig:self}
\end{figure}

The overall training pipeline of our proposed method, EpipolarPose, is given in Figure \ref{fig:self}. The orange-background part shows the inference pipeline. For training of EpipolarPose, the setup is assumed to be as follows. There are $n$ cameras ($n \geq 2$ must hold) which simultaneously take the picture of the person in the scene. The cameras are given id numbers from $1$ to $n$ where consecutive cameras are close to each other (\ie they have small baseline). The cameras produce images $I_1$, $I_2$, \dots $I_n$. Then, the set of consecutive image pairs, $\{(I_i, I_{i+1}) | i=1,2,\dots,n-1\}$, form the training examples.

\subsection{Training} 
In the training pipeline of EpipolarPose (Figure \ref{fig:self}), there are two branches each starting with the same pose estimation network (a ResNet followed by a deconvolution network \cite{Sun_2018_ECCV}). These networks were pre-trained on the MPII Human Pose dataset (MPII) \cite{mpii}. During training, only the pose estimation network in the upper branch is trained; the other one is kept frozen. 

EpipolarPose can be trained using more than $2$ cameras but for the sake of simplicity, here we will describe the training pipeline for $n=2$.  For $n=2$, each training example contains only one image pair. Images $I_i$ and $I_{i+1}$ are fed into both the 3D (upper) branch and 2D (lower) branch pose estimation networks to obtain volumetric heatmaps $\hat{H}, H \in \mathbb{R}^{w \times h \times d}$ respectively, where $w, h$ are the spatial size after deconvolution, $d$ is the depth resolution defined as a hyperparameter. After applying soft argmax activation function $\varphi(\cdot)$ we get 3D pose $\hat{V} \in \mathbb{R}^{J \times 3}$ and 2D pose $U \in \mathbb{R}^{J \times 2}$ outputs where $J$ is the number of body joints. From a given  volumetric heatmap, one can obtain both a 3D pose (by applying softargmax to all 3 dimensions) and a 2D pose (by applying softargmax to only $x,y$). 

As an output of 2D pose branch, we want to obtain the 3D human pose $V$ in the global coordinate frame. Let the 2D coordinate of the $j^{th}$  joint in the $i^{th}$ image be $U_{i,j} = [x_{i,j}, y_{i,j}]$ and its 3D coordinate be $[ X_j, Y_j, Z_j ]$, we can describe the relation between them assuming  a pinhole image projection model
\begin{equation}\small\begin{aligned}
    \begin{bmatrix} x_{i,j}\\ y_{i,j}\\ w_{i,j} \end{bmatrix}
    = K \left[R|RT\right] \begin{bmatrix} X_j\\ Y_j\\ Z_j \\ 1 \end{bmatrix},
    K = \begin{bmatrix} f_x &0 &c_x\\ 0 &f_y &c_y\\ 0 &0 &1 \end{bmatrix},
    T = \begin{bmatrix} T_x \\ T_y\\ T_z \end{bmatrix},
\end{aligned}\label{eq:camera}\end{equation}
where $w_{i,j}$ is the depth of the $j^{th}$  joint in the $i^{th}$ camera’s image with respect to the camera reference frame, $K$ encodes the camera intrinsic parameters (\eg, focal length $f_x$ and $f_y$, principal point $c_x$ and $x_y$), $R$ and $T$ are camera extrinsic parameters of rotation and translation, respectively. We omit camera distortion for simplicity.

When camera extrinsic parameters are not available, which is usually the case in dynamic capture environments, we can use body joints as calibration targets. We assume the first camera as the center of the coordinate system, which means $R$ of the first camera is identity. For corresponding joints in  $U_i$ and  $U_{i+1}$, in the image plane, we find the fundamental matrix $F$ satisfying $U_{i,j} F U_{i+1,j} = 0$ for $\forall j$ using the RANSAC algorithm. From $F$, we calculate the essential matrix $E$ by $E = K^T F K$. By decomposing $E$ with SVD, we obtain 4 possible solutions to $R$. We decide on the correct one by verifying possible pose hypotheses by doing cheirality check. The cheirality check basically means that the triangulated 3D points should have positive depth \cite{Nister2004}. We omit the scale during training, since our model uses normalized poses as ground truth.

Finally, to obtain a 3D pose $V$ for corresponding synchronized 2D images, we utilize triangulation (\ie epipolar geometry) as follows. For all joints in $(I_i, I_{i+1})$ that are not occluded in either image, triangulate a 3D point $[X_j, Y_j, Z_j]$ using polynomial triangulation \cite{Hartley1997}. For settings including more than 2 cameras, we calculate the vector-median to find the median 3D position.

To calculate the loss between 3D pose in camera frame $\hat{V}$ predicted by the upper (3D) branch, we project $V$ onto corresponding camera space, then minimize  $\mathrm{smooth}_{L_1}(V-\hat{V})$ to train the 3D branch where 

\begin{equation}
\mathrm{smooth}_{L_1}(x) = \begin{cases} 0.5x^2 & \text{if } |x| < 1 \\ |x| - 0.5 & \text{otherwise} \end{cases}
\label{eq:loss}
\end{equation}

\paragraph{Why do we need a frozen 2D pose estimator?} In the training pipeline of EpipolarPose, there are two branches each of which is starting with a pose estimator. While the estimator in the upper branch is trainable,  the other one in the lower branch is frozen. The job of the lower branch estimator is to produce 2D poses. One might question the necessity of the frozen estimator since we could obtain 2D poses from the trainable upper branch as well. When we tried to do so, our method produced degenerate solutions where all keypoints collapse to a single location. In fact, other multi-view methods faced the same problem \cite{rhodin2018,suwajanakorn}.  Rhodin \etal \cite{rhodin2018} solved this problem by using a small set of ground-truth  examples, however, obtaining such  ground-truth may not be feasible in most of the in the wild settings. Another solution proposed recently \cite{suwajanakorn} is to minimize angular distance between estimated relative rotation $\hat{R}$ (computed via Procrustes alignment of the two sets of keypoints) and the ground truth $R$. Nevertheless, it is hard to obtain ground truth $R$ in dynamic capture setups. To overcome these shortcomings, we utilize a frozen 2D pose detector during training time only.
\subsection{Inference}

Inference involves the orange-background part in Figure \ref{fig:self}. The input is just a single image and the output is the estimated 3D pose $\hat{V}$ obtained by a soft-argmax activation, $\varphi(\cdot)$, on 3D volumetric heatmap $\hat{H}_i$.

\begin{figure}
\centering
\includegraphics[width=\columnwidth]{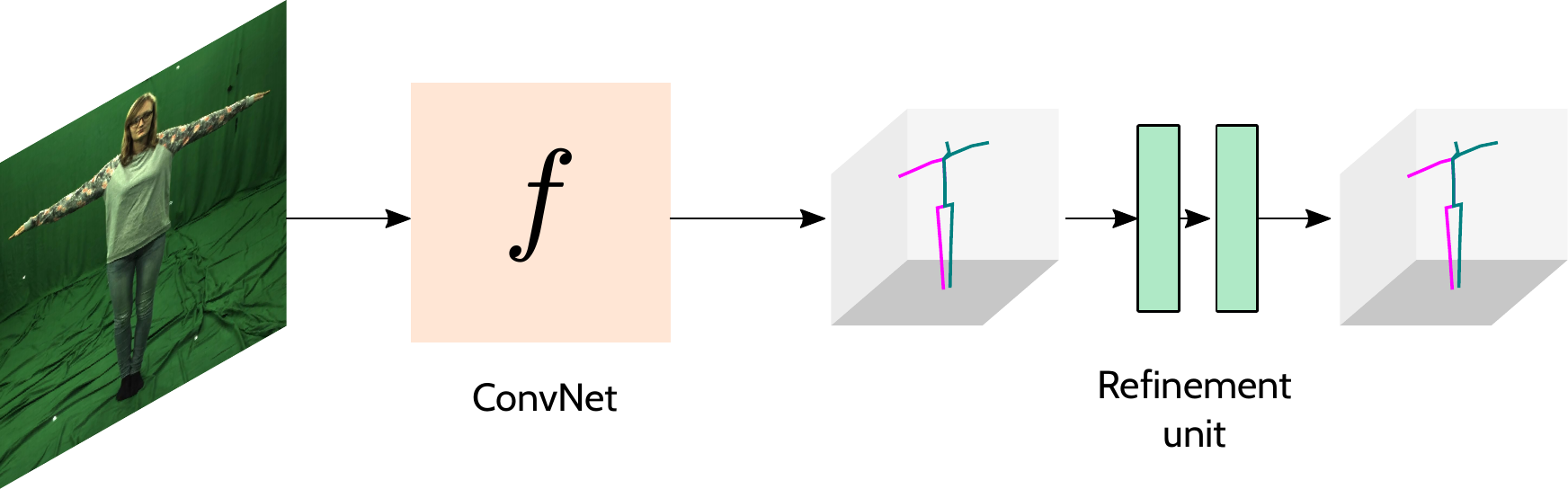}
\caption{Overall inference pipeline with a refinement unit which is an optional stage to refine the predictions of the model trained with self supervision. The $f$ function denotes the inference function (orange-background part in Figure \ref{fig:self}) of EpipolarPose. }
\label{fig:sum}
\end{figure}
\subsection{Refinement, an optional post-training} 

In the literature there are several techniques \cite{martinez_2017_3dbaseline,posegrammar2018, fuse_tekin} to lift detected 2D keypoints into 3D joints. These methods are capable of learning generalized 2D$\rightarrow$3D mapping which can be obtained from motion capture (MoCap) data by simulating random camera projections. Integrating a refinement unit (RU) to our self supervised model can further improve the pose estimation accuracy. In this way, one can train EpipolarPose on his/her own data which consists of multiple view footages without any labels and integrate it with RU to further improve the results. To make this possible, we modify the input layer of RU to accept noisy 3D detections from EpipolarPose and make it learn a refinement strategy. (See Figure \ref{fig:sum})

The overall RU architecture is inspired by \cite{martinez_2017_3dbaseline,posegrammar2018}. It has 2 computation blocks which have certain linear layers followed by Batch Normalization \cite{batchnormalization}, Leaky ReLU \cite{leakyrelu} activation and Dropout layers to map 3D noisy inputs to more reliable 3D pose predictions. To facilitate information flow between layers, we add residual connections \cite{He2016} and apply intermediate loss to expedite the intermediate layers’ access to supervision.

\begin{figure}
\centering
\includegraphics[width=\columnwidth]{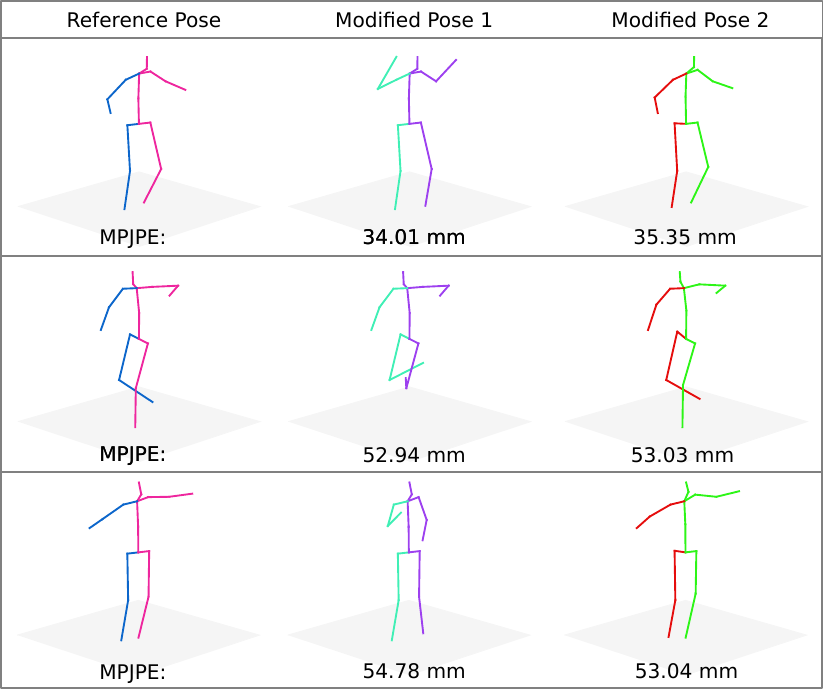}
\caption{\textbf{Left}: reference poses from Human3.6M dataset. \textbf{Middle}: manually modified poses to obtain similar MPJPE with poses on the \textit{right}, yet structured differently from reference poses. \textbf{Right}: poses obtained by adding random gaussian noise to each body joint.}
\label{fig:pose}
\end{figure}
\subsection{Pose Structure Score}
\label{sec:pss}

As we discussed in Section \ref{sec:intro}, traditional distance-based evaluation metrics (such as MPJPE, PCK) treat each joint independently, hence, fail to asses the whole pose as a structure. In Figure \ref{fig:pose}, we present example poses that have the same MPJPE but are structurally very different, with respect to a reference pose.  

We propose a new performance measure, called the Pose Structure Score (PSS), which is sensitive to structural errors in pose. PSS computes a scale invariant performance score with the capability to assess the structural plausibility of a pose with respect to its ground truth. Note that PSS is not a loss function, it is a performance score that can be used along with MPJPE and PCK to account for structural errors made by the  pose estimator. PSS is an indicator about the deviation from the ground truth pose that has the potential to cause a wrong inference in a subsequent task requiring semantically meaningful poses, \eg action recognition, human-robot interaction.

\paragraph{How to compute PSS?} The computation of PSS requires a reference distribution of ground-truth poses. Given a ground-truth set composed of $n$ poses $\mathbf{q}_i, i \in \{ 1, \cdots, n\}$, we normalize each pose vector as $\hat{ \mathbf{q} }_i = \frac{\mathbf{q}_i} {||\mathbf{q}_i||}$. Then, we compute  $k$ cluster centers $\boldsymbol{\mu}_j, j \in \{1,\cdots,k\}$ using $k$-means clustering.
Then, to compute the PSS of a predicted pose $\mathbf{p}$ against its ground-truth pose $ \mathbf{q}$, we use 

\begin{gather}
\label{eq:pss}
\mathrm{PSS}(\mathbf{p}, \mathbf{q}) = \delta \big (C(\mathbf{p}), C(\mathbf{q}) \big )\;\;\;\;\mathrm{where} 
\\
C(\mathbf{p}) = \underset{k}{\arg \min}{||\mathbf{p}-\boldsymbol{\mu}_k||_2^2},  \;\;\;\;\; \delta(i,j) = \begin{cases} 1 & i = j \\ 0 & i \ne j \end{cases}
\end{gather}

\begin{figure}
\centering
\includegraphics[width=\columnwidth]{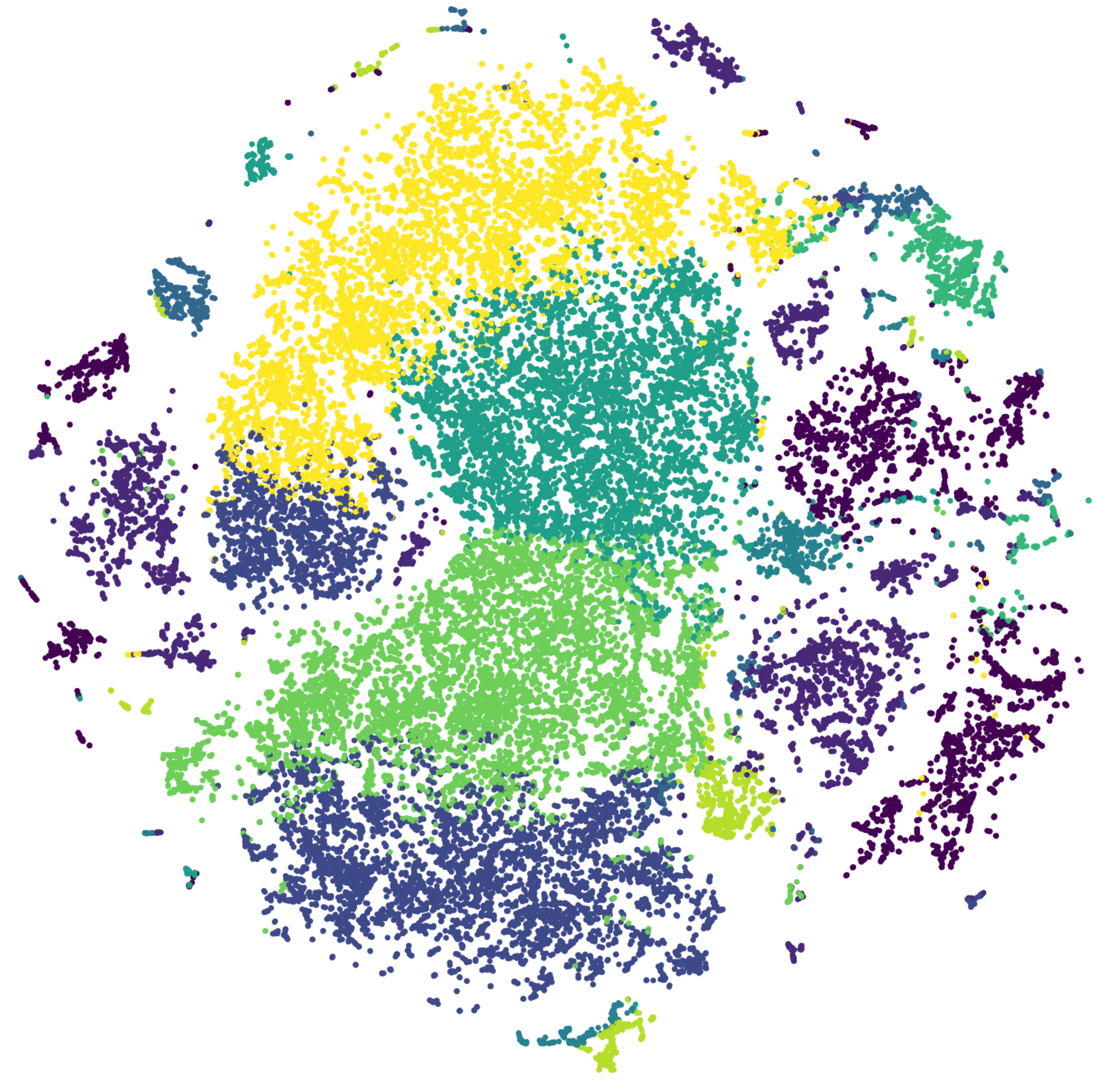}
\caption{\textbf{t-SNE graph of human poses after clustering.} Here we choose $k=10$ for visualization purposes. Each color represents a cluster.}
\label{fig:tsne}
\end{figure}

\begin{figure}
\centering
\includegraphics[width=\columnwidth]{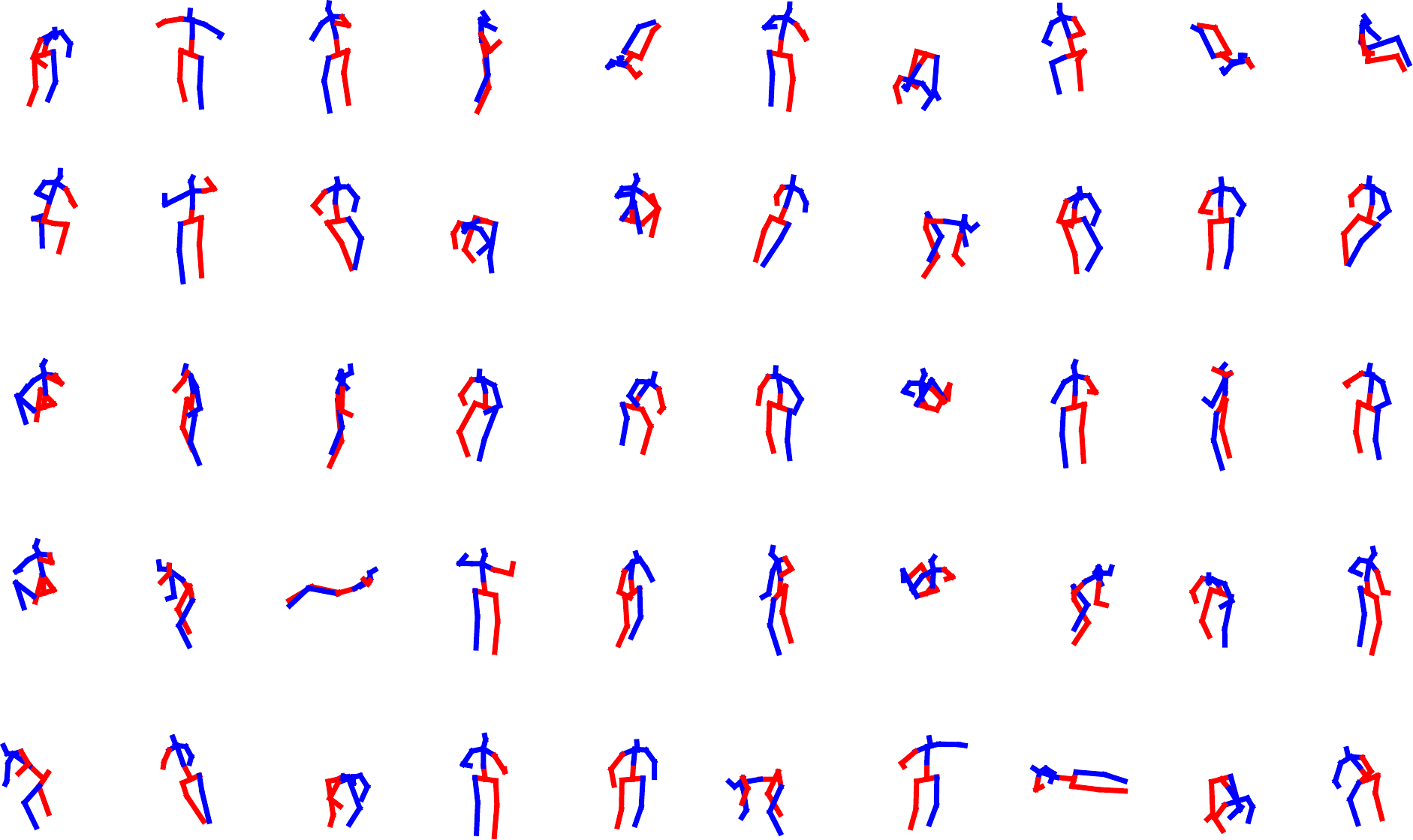}
\caption{\textbf{Cluster centers} which represents the canonical poses in Human3.6M ($k=50$).}
\label{fig:atom}
\end{figure}

The mPSS, mean-PSS, of a set of poses is the average over their individual scores as computed in Eq. \eqref{eq:pss}. Figure \ref{fig:tsne} shows the t-SNE \cite{tsne} graph of poses and clusters. Figure \ref{fig:atom} depicts the cluster centers which represent canonical poses.

In our experiments, we chose the number of pose clusters as 50 and 100. We denoted the corresponding PSS results with mPSS@50 and mPSS@100 expressions. Note that mPSS gives the percentage of structurally correct poses, therefore higher scores are better. To test the stability of our clustering, we ran $k$-means 100 times each with random initializations. Then, for each pair of runs, we established pairwise correspondences between clusters. For each correspondence, we computed intersection over union (IOU). Average IOU over all pairings and correspondences turned out to be $0.78$. Additionally, the mPSS of different pose estimation models vary $\pm 0.1\%$ when we use different $k$-means outputs as reference. These analyses prove the stability of PSS.

\paragraph{Implementation details}
We use the Integral Pose \cite{Sun_2018_ECCV} architecture for both 2D and 3D branches with a ResNet-50 \cite{He2016} backend. Input image and output heatmap sizes are $256 \times 256$ and $J  \times 64 \times 64 \times 64$, respectively where $J$ is the number of joints. We initialize all models used in experiments after training on the MPII \cite{mpii}.

During training, we use mini-batches of size 32, each one containing $I_i, I_{i+1}$ image pairs. If more than two cameras are available, we include the views from all cameras in a mini-batch. We train the network for 140 epochs using Adam optimizer \cite{adam} with a learning rate of $10^{-3}$ multiplied with $0.1$ at steps $90$ and $120$. Training data is augmented by random rotations of $\pm 30^{\circ}$ and scaled by a factor between $0.8$ and $1.2$. Additionally, we utilize synthetic occlusions \cite{Sarandi18IROSW} to make the network robust to occluded joints. For the sake of simplicity, we run the 2D branch once to produce triangulated 3D targets and train the 3D branch using cached labels. We implemented the whole pipeline using PyTorch \cite{pytorch}.
\section{Experiments}
\paragraph{Datasets.}
We first conduct experiments on the Human3.6M (H36M) large scale 3D human pose estimation benchmark \cite{h36m}. It is one of the largest datasets for 3D human pose estimation with 3.6 million images featuring 11 actors performing 15 daily activities, such as eating, sitting, walking and taking a photo, from 4 camera views. We mainly use this dataset for both quantitative and qualitative evaluation.

We follow the standard protocol on H36M and use the subjects 1, 5, 6, 7, 8 for training and the subjects 9, 11 for evaluation. Evaluation is performed on every $64^{th}$ frame of the test set. We include average errors for each method.

To demonstrate further applicability of our method, we use MPI-INF-3DHP (3DHP) \cite{monocular3d} which is a recent dataset that includes both indoor and outdoor scenes. We follow the standard protocol: The five chest-height cameras and the provided 17 joints (compatible with H36M) are used for training. For evaluation, we use the official test set which includes challenging outdoor scenes. We report the results in terms of PCK and NPCK to be consistent with \cite{rhodin2018}. Note that we do not utilize any kind of background augmentation to boost the performance for outdoor test scenes.

\begin{table}
\caption{\textbf{Triangulation results on H36M.} Effects of different 2D keypoint sources on triangulation performance. \textit{GT 2D} denotes the usage of ground truth 2D labels. \textit{H36M 2D} and \textit{MPII 2D} shows the pose estimation models trained on those datasets.}

\label{table:tri}

\resizebox{1\columnwidth}{!}{
\begin{tabular}{|l||c|c|c|c|c|}
\multicolumn{6}{c}{}\\
\hline
\bf Methods & \bf  MPJPE & \bf NMPJPE & \bf PMPJPE & \bf mPSS@50 & \bf mPSS@100 \\
\hline
\hline
Pavlakos \etal \cite{pavlakos2017harvesting} & 56.89 & - & - & - & - \\
\hline
GT 2D & 4.38 & 2.87 & 2.13 & 98.93 & 97.16 \\
GT 2D (w/o $\mathbf{R}$) & n/a & 22.46 & 15.06 & 98.83 & 96.03 \\
H36M 2D & 28.37 & 26.28 & 25.19 & 95.08 & 94.2 \\
MPII 2D & 45.86 & 37.79 & 36.83 & 90.06 & 85.96 \\ 
\hline
\end{tabular}
}

\end{table}
\paragraph{Metrics.}
We evaluate pose accuracy in terms of MPJPE (mean per joint position error), PMPJPE (procrustes aligned mean per joint position error), PCK (percentage of correct keypoints), and PSS at scales @50 and @100. To compare our model with \cite{rhodin2018}, we measured the normalized metrics NMPJPE and NPCK, please refer to \cite{rhodin2018} for further details. Note that PSS, by default, uses normalized poses during evaluation. In the presented results “n/a” means “not applicable” where it’s not possible to measure respective metric with provided information, “-” means “not available”. For instance, it’s not possible to measure MPJPE or PCK when $R$, the camera rotation matrix, is not available. For some of the previous methods with open source code, we indicate their respective PSS scores. We hope, in the future, PSS will be adapted as an additional performance measure, thus more results will become available for complete comparisons. 

\begin{table*}[]
\centering
\caption{\textbf{H36M results.} \textbf{Top:} Comparison of results between our methods trained with different settings and the state-of-the-art fully supervised methods. (\textit{FS}: fully supervised, \textit{SS}: self supervised) \textbf{Bottom:} Effect of adding refinement unit (RU) over SS. (* uses the 2D keypoints from an MPII pre trained model as input, hence is comparable to our SS+RU model.)}
\label{table:sota}
\resizebox{0.80\textwidth}{!}{

\begin{tabular}{|l||c|c|c|c|c|}
\multicolumn{6}{c}{Supervised training on all subjects of H36M}\\
\hline
\bf Methods & \bf MPJPE & \bf NMPJPE & \bf PMPJPE & \bf mPSS@50 & \bf mPSS@100 \\
\hline
\hline

Nie \etal \cite{Nie_2017_ICCV} (ICCV’17) &97.5 &- &79.5 &- &- \\
Sanzari \etal \cite{sanzari} (ECCV'16)& 93.1 & - & - & - & - \\    
Tome \etal \cite{Tome_2017_CVPR} (CVPR'17) & 88.4 & - & - & 73.0 &  58.8 \\    
Rogez \etal \cite{Rogez_2017_CVPR} (CVPR'17) & 87.7 & -  & 71.6 &- &- \\
Pavlakos \etal \cite{pavlakos17volumetric} (CVPR'17) &71.9&-&-&74.05&53.93 \\ 
Rhodin \etal \cite{rhodin2018} (CVPR'18)& 66.8 & 63.3 & 51.6 & - & - \\
Martinez \etal \cite{martinez_2017_3dbaseline} (ICCV'17) & 62.9 & - & 47.7 & 78.12 & 73.26 \\
Pavlakos \etal \cite{pavlakos2018ordinal} (CVPR'18)& 56.2 & - & - & 80.03 & 69.18 \\
Sun \etal \cite{Sun_2018_ECCV} (ECCV'18)& 49.6 & - & 40.6 & - & - \\ 

\hline
Ours FS & 51.83 & 51.58 & 45.04 & 84.44 & 78.67 \\
Ours SS & 76.60 & 75.25 & 67.45 & 73.09 & 64.03 \\
Ours SS (w/o $\mathbf{R}$) & n/a & 77.75 & 70.67 & 70.67 & 62.05 \\ 
\hline
\multicolumn{6}{c}{}\\
\multicolumn{6}{c}{Integrating Refinement Unit with SS trained network on H36M}\\
\hline
\bf Methods & \bf MPJPE & \bf NMPJPE & \bf PMPJPE & \bf mPSS@50 & \bf mPSS@100 \\
\hline
\hline
Martinez \etal \cite{martinez_2017_3dbaseline} (ICCV'2017)* & 67.5 & - & 52.5 & - & - \\
Ours SS + RU & \bf 60.56 & \bf 60.04 & \bf 47.48 & \bf 80.42 & \bf 75.41 \\
\hline
\end{tabular}

}
\end{table*} 
\subsection{Results}
\subsection*{Can we rely on the labels from multi view images?}
Table \ref{table:tri} summarizes triangulation results from different 2D keypoint sources on the H36M dataset. Note that we use training subjects to obtain these results, since our goal is to find out  the performance of triangulation on the training data. Overall, the quality of estimated keypoints is crucial to attain better results. If we have the ground truth 2D keypoints and camera geometry, triangulation gives 4.3 mm error and 99\% PSS which is near perfect. Lack of camera geometry reduces the PMPJE and mPSS@50 by a small amount of 13 mm and 1\%, respectively. A pose detector trained on the 2D labels of H36M improves the MPII-pretrained one up to 17 mm and 5\%. Note that, it is expected to have slightly worse performance when evaluating the MPII-pretrained detector on the H36M validation set. Data in H36M was captured with markers, and therefore, have high accuracy and consistency in 2D annotations across subject and scenes; on the other hand, the annotations in MPII were done by humans and some of the keypoints are localized differently. For instance, shoulders and hips are closer to edges of the body in the MPII dataset. 

Compared  to Pavlakos \etal’s \cite{pavlakos2017harvesting} results, our triangulation using an MPII-pretrained detector is 11mm better in terms of MPJPE.
\subsection*{Comparison to State-of-the-art}
In Table \ref{table:sota}, we present the results of our model with different supervision types in comparison with recent state-of-the-art methods. We present the fully supervised (FS) version of our model to provide a baseline. Our own implementation of  “Integral Pose” architecture \cite{Sun_2018_ECCV} produced a slightly different result than reported. The difference between our result (52mm) and the reported one \cite{Sun_2018_ECCV} (49mm) can be attributed to the authors’ 2D-3D mixed training which we refrained from doing in order to decouple 3D pose estimation stage from 2D.

Our self supervised (SS) model performs quite well compared to the recent fully 3D supervised methods \cite{pavlakos17volumetric, Rogez_2017_CVPR, sanzari, Tome_2017_CVPR} which require abundant labeled data to learn. Obtaining comparable results to state-of-the-art methods without using any 3D ground truth examples is a promising step for such a nontrivial task.

Refinement Unit (RU) which is an optional extension to our SS network is helpful for achieving better results. Adding RU further improves the performance of our SS model by 20\% . To measure the representation capacity of the outputs from our SS model, we compare its result with Martinez \etal’s work \cite{martinez_2017_3dbaseline}. Since the RU architecture is identical to Martinez \etal, we selected their model trained with 2D keypoints from an MPII-pretrained pose detector for a fair comparison. This results show that 3D depth information learned by our SS training method provides helpful cues to improve the performance of 2D-3D lifting approaches.

In Table \ref{table:self_mpi} \textit{top}, we show the FS training results on the 3DHP dataset as a baseline. We further use that information to analyze the differences between FS and SS training.   
\subsection*{Weakly/Self Supervised Methods}
Table \ref{table:self_h36m} outlines the performance of weakly/self supervised methods in the literature along with ours on the H36M dataset. The top part includes the methods not requiring paired 3D supervision. Since Tung \etal \cite{tung2017} use unpaired 3D ground truth labels that are easier to obtain, we place them here. Our SS model (with or without $R$) outperforms all previous methods \cite{tung2017,pavlakos2017harvesting} by a large margin in MPJPE metric. We observe a large difference (21mm) between training with ground truth 2D triangulations and MPII-pretrained ones. This gap indicates us that the 2D keypoint estimation quality is crucial for better performance.

To better understand the source of performance gain in ours and Rhodin \etal, we can analyze the gap between the models trained with full supervision (FS) and subject 1 of H36M and 3DHP only (S1). In our method, the difference between FS and S1 training is 12 and 9mm, while Rhodin \etal’s difference is 15 and 18mm for H36M and 3DHP, respectively (lower is better). It shows us that our learning strategy is better at closing the gap. Even though Rhodin \etal uses S1 for training, our SS method outperforms it on H36M dataset. In the case of S1 training, there is an explicit improvement (14mm, 4mm for H36M and 3DHP) with our approach. Also, SS training with our method on 3DHP has comparable results to Rhodin \etal’ S1. 

Finally, the \textit{bottom} part in Table \ref{table:self_h36m} gives a fair comparison of our model against Drover \etal’s since they report results only with 14 joints. Our method yields 4mm less error than their approach.

\begin{table*}[h]
\centering
\caption{\textbf{H36M weakly/self supervised results.} \textbf{Top:} Methods that can be trained without 3D ground truth labels. (Tung \etal \cite{tung2017} uses unpaired 3D supervision which is easier to get. \textit{3DInterp} denotes the results of \cite{3dinterpreter} implemented by \cite{tung2017}. \textit{2D GT} denotes training with triangulations obtained from ground truth 2D labels.) \textbf{Middle:} Methods requiring a small set of ground truth data. (S1 denotes using ground truth labels of H36M subject \#1 during training.) \textbf{Bottom:} Comparison to Drover \etal \cite{drover2018} that evaluated using 14 joints (14j)}
\label{table:self_h36m}
\resizebox{0.88\textwidth}{!}{

\begin{tabular}{|l||c|c|c|c|c|}
\multicolumn{6}{c}{Training without ground truth data}\\
\hline
\bf Methods & \bf  MPJPE & \bf NMPJPE & \bf PMPJPE & \bf mPSS@50 & \bf mPSS@100 \\
\hline
\hline
Pavlakos \etal  \cite{pavlakos2017harvesting} (CVPR'2017) & 118.41 & - & - & -& - \\
Tung \etal - 3DInterp \cite{tung2017} (ICCV'2017) & 98.4 & - & - & - & - \\
Tung \etal \cite{tung2017} (ICCV'2017) & 97.2 & - & - & - & - \\
\hline
Ours SS & 76.60 & 75.25 & 67.45 & 73.09 & 64.03 \\
Ours SS (w/o $\mathbf{R}$) & n/a & 77.75 & 70.67 & 70.67 & 62.05 \\
\hline
Ours SS (2D GT) & 55.08 & 54.90 & 47.91 & 83.9 & 78.69 \\
\hline
\multicolumn{6}{c}{}\\
\multicolumn{6}{c}{Training with only Subject 1 of H36M}\\

\hline
\bf Methods & \bf  MPJPE & \bf NMPJPE & \bf PMPJPE & \bf mPSS@50 & \bf mPSS@100 \\
\hline
\hline

Rhodin \etal \cite{rhodin2018} S1 & n/a & 78.2 & 64.6 & - & - \\
Rhodin \etal \cite{rhodin2018} S1 (w/o $\mathbf{R})$ & n/a & 80.1 & 65.1 & - & - \\
\hline
Ours S1 & \bf 65.35 & \bf 64.76 & \bf 57.22 & \bf 81.91 & \bf 75.2 \\
Ours S1 (w/o $\mathbf{R})$ & n/a & 66.98 & 60.16 & 77.65 & 72.4 \\
\hline
\multicolumn{6}{c}{}\\
\multicolumn{6}{c}{Evaluation using 14 joints}\\

\hline
\bf Methods & \bf  MPJPE & \bf NMPJPE & \bf PMPJPE & \bf mPSS@50 & \bf mPSS@100 \\
\hline
\hline

Drover \etal \cite{drover2018}(14j) (ECCVW'2018) & - & - & 64.6 & - & - \\
Ours SS (14j) & \bf 69.94 & \bf 67.90 & \bf 60.24 & n/a & n/a \\
\hline
\end{tabular} 

}
\end{table*}

\begin{table*}[h]
\centering
\caption{\textbf{3DHP results.} \textbf{Top:} Fully supervised training results. \textbf{Middle:} Self supervised learning using only subject 1. \textbf{Bottom:} Self supervised training without any ground truth examples.}
\label{table:self_mpi}
\resizebox{0.88\textwidth}{!}{

\begin{tabular}{|l||c|c|c|c|c|c|}
\multicolumn{6}{c}{Supervised training}\\
\hline
\bf Methods & \bf MPJPE & \bf NMPJPE & \bf PCK & \bf NPCK & \bf mPSS@50 & \bf mPSS@100 \\
\hline
\hline
Mehta \etal \cite{monocular3d} & - & - & 72.5 & - & - & - \\
Rhodin \etal \cite{rhodin2018} FS & n/a & \bf 101.5 & n/a & \bf 78.8 & - & - \\
Ours FS & \bf 108.99 & 106.38 & \bf 77.5 & 78.1 & 87.15 & 82.21 \\
\hline

\multicolumn{6}{c}{}\\
\multicolumn{6}{c}{Training with only Subject 1 of 3DHP}\\
\hline
\bf Methods & \bf MPJPE & \bf NMPJPE & \bf PCK & \bf NPCK & \bf mPSS@50 & \bf mPSS@100 \\
\hline
\hline

Rhodin \etal \cite{rhodin2018} S1 & n/a & 119.8 & n/a & 73.1 & - & - \\
Rhodin \etal \cite{rhodin2018} S1 (w/o $\mathbf{R}$) & n/a & 121.8 & n/a & 72.7 & - & - \\

\hline
Ours S1 & n/a & \bf 115.37 & n/a & \bf 74.4 & \bf 75.64 & \bf 73.15 \\
Ours S1 (w/o $\mathbf{R}$) & n/a & 119.86 & n/a & 73.5 & 73.41 & 70.97 \\
\hline

\multicolumn{6}{c}{}\\
\multicolumn{6}{c}{Training without ground truth data}\\
\hline
\bf Methods & \bf MPJPE & \bf NMPJPE & \bf PCK & \bf NPCK & \bf mPSS@50 & \bf mPSS@100 \\
\hline
\hline
Ours SS & 126.79 & 125.65 & 64.7 & 71.9 & 70.94 & 67.58 \\
\hline
\end{tabular} 
}

\end{table*}

\begin{figure}[h]
\centering
\includegraphics[width=\columnwidth]{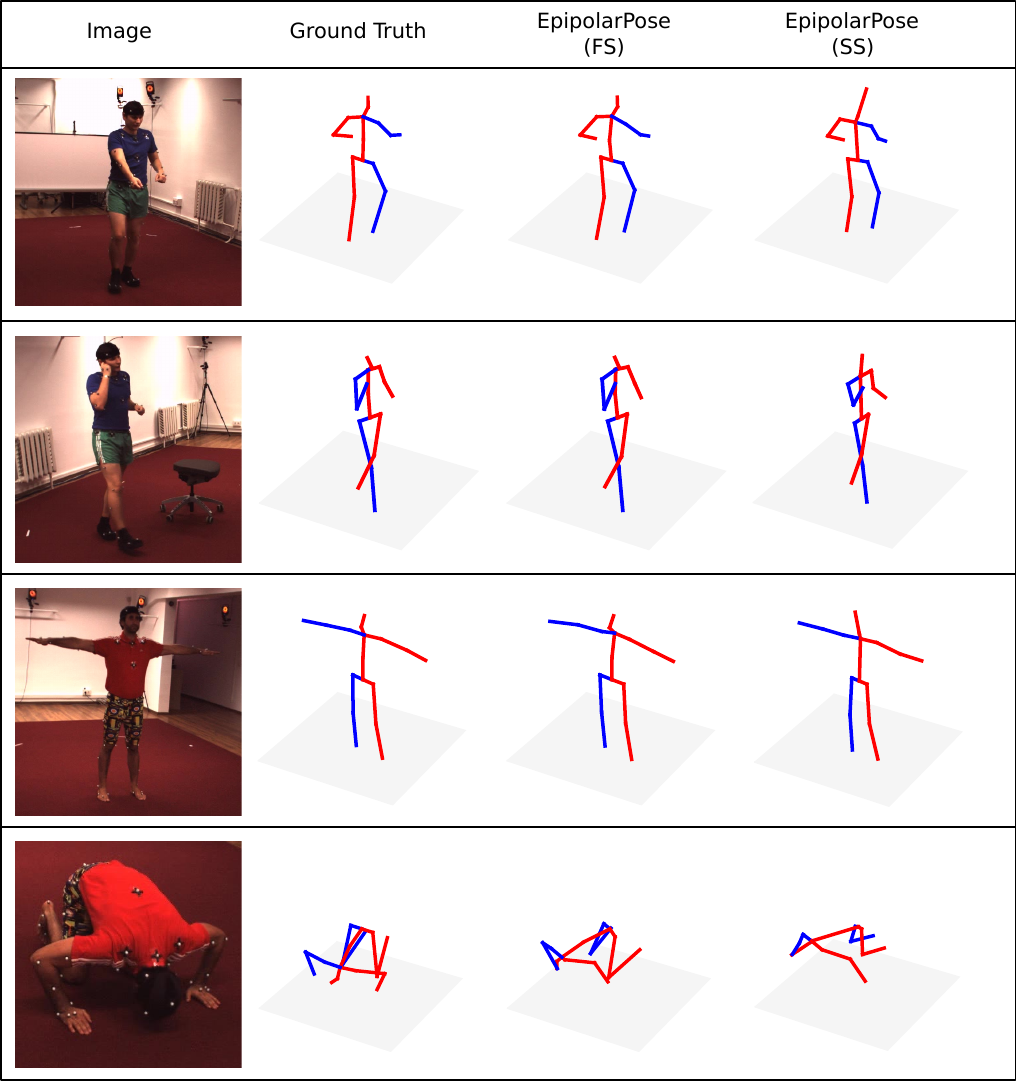}
\caption{\textbf{Qualitative results on H36M dataset.} Provided 3D poses are from different camera views for better visualization. Last row depicts a failure case. (\textit{FS:} fully supervised training, \textit{SS:} self supervised training)}
\label{fig:qual}
\end{figure}

\section{Conclusion}
In this work, we have shown that even without any 3D ground truth data and the knowledge of camera extrinsics, multi view images can be leveraged to obtain self supervision. At the core of our approach, there is EpipolarPose which can utilize 2D poses from multi-view images using epipolar geometry to self-supervise a 3D pose estimator. EpipolarPose achieved state-of-the-art results in Human3.6M and MPI-INF-3D-HP benchmarks among weakly/self-supervised methods. In addition, we discussed the weaknesses of localization based metrics \ie MPJPE and PCK for human pose estimation task and therefore proposed a new performance measure Pose Structure Score (PSS) to score the structural plausibility of a pose with respect to its ground truth.

{\small
\bibliographystyle{ieee_fullname}
\bibliography{references}

\begin{thebibliography}{10}\itemsep=-1pt

\bibitem{amin2013}
Sikandar Amin, Mykhaylo Andriluka, Marcus Rohrbach, and Bernt Schiele.
\newblock Multi-view pictorial structures for 3d human pose estimation.
\newblock In {\em British Machine Vision Conference}, 2013.

\bibitem{mpii}
Mykhaylo Andriluka, Leonid Pishchulin, Peter Gehler, and Bernt Schiele.
\newblock 2{D} human pose estimation: New benchmark and state of the art
  analysis.
\newblock In {\em IEEE Conference on Computer Vision and Pattern Recognition},
  2014.

\bibitem{belagiannis2014}
Vasileios Belagiannis, Sikandar Amin, Mykhaylo Andriluka, Bernt Schiele, Nassir
  Navab, and Slobodan Ilic.
\newblock 3{D} pictorial structures for multiple human pose estimation.
\newblock In {\em IEEE Conference on Computer Vision and Pattern Recognition},
  2014.

\bibitem{belagiannis2016}
Vasileios Belagiannis, Sikandar Amin, Mykhaylo Andriluka, Bernt Schiele, Nassir
  Navab, and Slobodan Ilic.
\newblock 3{D} pictorial structures revisited: Multiple human pose estimation.
\newblock {\em {IEEE} Transaction on Pattern Analysis and Machine
  Intelligence}, 2016.

\bibitem{martin2010}
Martin Bergtholdt, Jorg Kappes, Stefan Schmidt, and Christoph Schnorr.
\newblock A study of parts-based object class detection using complete graphs.
\newblock In {\em International Journal of Computer Vision}, 2010.

\bibitem{burenius2013}
Magnus Burenius, Josephine Sullivan, and Stefan Carlsson.
\newblock 3{D} pictorial structures for multiple view articulated pose
  estimation.
\newblock In {\em IEEE Conference on Computer Vision and Pattern Recognition},
  2013.

\bibitem{cao2017}
Zhe Cao, Tomas Simon, Shih-En Wei, and Yaser Sheikh.
\newblock Realtime multi-person 2d pose estimation using part affinity fields.
\newblock In {\em IEEE Conference on Computer Vision and Pattern Recognition},
  2017.

\bibitem{2dposematching}
Ching-Hang Chen and Deva Ramanan.
\newblock 3{D} human pose estimation = 2{D} pose estimation + matching.
\newblock In {\em IEEE Conference on Computer Vision and Pattern Recognition},
  2017.

\bibitem{drover2018}
Dylan Drover, Rohith MV, Ching-Hang Chen, Amit Agrawal, Ambrish Tyagi, and
  Cong~Phuoc Huynh.
\newblock Can 3d pose be learned from 2d projections alone?
\newblock {\em European Conference on Computer Vision Workshops}, 2018.

\bibitem{elhayek2017}
Ahmed Elhayek, Edilson de Aguiar, Arjun Jain, Jonathan Thompson, Leonid
  Pishchulin, Micha Andriluka, Christoph Bregler, Bernt Schiele, and Christian
  Theobalt.
\newblock {MARCOnI}{\textemdash}{ConvNet}-based {MARker}-less motion capture in
  outdoor and indoor scenes.
\newblock {\em {IEEE} Transaction on Pattern Analysis and Machine
  Intelligence}, 2017.

\bibitem{posegrammar2018}
Hao-Shu Fang, Yuanlu Xu, Wenguan Wang, Xiaobai Liu, and Song-Chun Zhu.
\newblock Learning pose grammar to encode human body configuration for 3{D}
  pose estimation.
\newblock In {\em Association for the Advancement of Artificial Intelligence},
  2018.

\bibitem{Hartley1997}
Richard~I. Hartley and Peter Sturm.
\newblock Triangulation.
\newblock {\em Computer Vision and Image Understanding}, 1997.

\bibitem{He2016}
Kaiming He, Xiangyu Zhang, Shaoqing Ren, and Jian Sun.
\newblock Deep residual learning for image recognition.
\newblock In {\em IEEE Conference on Computer Vision and Pattern Recognition},
  2016.

\bibitem{batchnormalization}
Sergey Ioffe and Christian Szegedy.
\newblock Batch normalization: Accelerating deep network training by reducing
  internal covariate shift.
\newblock In {\em Journal of Machine Learning Research}, 2015.

\bibitem{h36m}
Catalin Ionescu, Dragos Papava, Vlad Olaru, and Cristian Sminchisescu.
\newblock Human3.6m: Large scale datasets and predictive methods for 3{D} human
  sensing in natural environments.
\newblock In {\em {IEEE} Transaction on Pattern Analysis and Machine
  Intelligence}, 2014.

\bibitem{adam}
Diederik~P. Kingma and Jimmy Ba.
\newblock Adam: A method for stochastic optimization.
\newblock In {\em International Conference on Learning Representations}, 2015.

\bibitem{kocabas2018}
Muhammed Kocabas, Salih Karagoz, and Emre Akbas.
\newblock Multiposenet: Fast multi-person pose estimation using pose residual
  network.
\newblock In {\em European Conference on Computer Vision}, 2018.

\bibitem{kulkarni}
Tejas~D Kulkarni, William~F Whitney, Pushmeet Kohli, and Josh Tenenbaum.
\newblock Deep convolutional inverse graphics network.
\newblock In {\em Advances in Neural Information Processing}, 2015.

\bibitem{monocular_asia}
Sijin Li and Antoni~B. Chan.
\newblock 3{D} human pose estimation from monocular images with deep
  convolutional neural network.
\newblock In {\em Asian Conference on Computer Vision}, 2014.

\bibitem{coco}
Tsung-Yi Lin, Michael Maire, Serge Belongie, James Hays, Pietro Perona, Deva
  Ramanan, Piotr Doll{\'{a}}r, and C.~Lawrence Zitnick.
\newblock Microsoft {COCO}: Common objects in context.
\newblock In {\em European Conference on Computer Vision}, 2014.

\bibitem{leakyrelu}
Andrew~L. Maas, Awni~Y. Hannun, and Andrew~Y. Ng.
\newblock Rectifier nonlinearities improve neural network acoustic models.
\newblock In {\em International Conference on Machine Learning}, 2013.

\bibitem{martinez_2017_3dbaseline}
Julieta Martinez, Rayat Hossain, Javier Romero, and James~J. Little.
\newblock A simple yet effective baseline for 3{D} human pose estimation.
\newblock In {\em International Conference on Computer Vision}, 2017.

\bibitem{monocular3d}
Dushyant Mehta, Helge Rhodin, Dan Casas, Pascal Fua, Oleksandr Sotnychenko,
  Weipeng Xu, and Christian Theobalt.
\newblock Monocular 3{D} human pose estimation in the wild using improve cnn
  supervision.
\newblock In {\em International Conference on 3DVision}, 2017.

\bibitem{Moreno_cvpr2017}
F. Moreno-Noguer.
\newblock 3{D} human pose estimation from a single image via distance matrix
  regression.
\newblock In {\em IEEE Conference on Computer Vision and Pattern Recognition},
  2017.

\bibitem{newell2016}
Alejandro Newell, Kaiyu Yang, and Jia Deng.
\newblock Stacked hourglass networks for human pose estimation.
\newblock In {\em European Conference on Computer Vision}, 2016.

\bibitem{Nister2004}
D. Nister.
\newblock An efficient solution to the five-point relative pose problem.
\newblock {\em {IEEE} Transaction on Pattern Analysis and Machine
  Intelligence}, 2004.

\bibitem{pytorch}
Adam Paszke, Sam Gross, Soumith Chintala, Gregory Chanan, Edward Yang, Zachary
  DeVito, Zeming Lin, Alban Desmaison, Luca Antiga, and Adam Lerer.
\newblock Automatic differentiation in pytorch.
\newblock In {\em International Conference on Learning Representations}, 2017.

\bibitem{pavlakos2018ordinal}
Georgios Pavlakos, Xiaowei Zhou, and Kostas Daniilidis.
\newblock Ordinal depth supervision for 3{D} human pose estimation.
\newblock In {\em IEEE Conference on Computer Vision and Pattern Recognition},
  2018.

\bibitem{pavlakos17volumetric}
Georgios Pavlakos, Xiaowei Zhou, Konstantinos~G Derpanis, and Kostas
  Daniilidis.
\newblock Coarse-to-fine volumetric prediction for single-image 3{D} human
  pose.
\newblock In {\em IEEE Conference on Computer Vision and Pattern Recognition},
  2017.

\bibitem{pavlakos2017harvesting}
Georgios Pavlakos, Xiaowei Zhou, Konstantinos~G Derpanis, and Kostas
  Daniilidis.
\newblock Harvesting multiple views for marker-less 3d human pose annotations.
\newblock In {\em IEEE Conference on Computer Vision and Pattern Recognition},
  2017.

\bibitem{rhodin2018}
Helge Rhodin, Jörg Spörri, Isinsu Katircioglu, Victor Constantin, Frédéric
  Meyer, Erich Müller, Mathieu Salzmann, and Pascal Fua.
\newblock Learning monocular 3d human pose estimation from multi-view images.
\newblock In {\em IEEE Conference on Computer Vision and Pattern Recognition},
  2018.

\bibitem{Rogez_2017_CVPR}
Gregory Rogez, Philippe Weinzaepfel, and Cordelia Schmid.
\newblock Lcr-net: Localization-classification-regression for human pose.
\newblock In {\em IEEE Conference on Computer Vision and Pattern Recognition},
  2017.

\bibitem{sanzari}
Marta Sanzari, Valsamis Ntouskos, and Fiora Pirri.
\newblock Bayesian image based 3d pose estimation.
\newblock In {\em European Conference on Computer Vision}, 2016.

\bibitem{Sarandi18IROSW}
Istv{\'a}n S{\'a}r{\'a}ndi, Timm Linder, Kai~O Arras, and Bastian Leibe.
\newblock How robust is 3d human pose estimation to occlusion?
\newblock In {\em IROS Workshop - Robotic Co-workers 4.0}, 2018.

\bibitem{compositional2017}
Xiao Sun, Jiaxiang Shang, Shuang Liang, and Yichen Wei.
\newblock Compositional human pose regression.
\newblock In {\em International Conference on Computer Vision}, 2017.

\bibitem{Sun_2018_ECCV}
Xiao Sun, Bin Xiao, Fangyin Wei, Shuang Liang, and Yichen Wei.
\newblock Integral human pose regression.
\newblock In {\em European Conference on Computer Vision}, 2018.

\bibitem{suwajanakorn}
Supasorn Suwajanakorn, Noah Snavely, Jonathan Tompson, and Mohammad Norouzi.
\newblock Discovery of latent 3d keypoints via end-to-end geometric reasoning.
\newblock In {\em Advances in Neural Information Processing}, 2018.

\bibitem{structured3d}
Bugra Tekin, Isinsu Katircioglu, Mathieu Salzmann, Vincent Lepetit, and Pascal
  Fua.
\newblock Structured prediction of 3{D} human pose with deep neural networks.
\newblock In {\em British Machine Vision Conference}, 2016.

\bibitem{fuse_tekin}
Bugra Tekin, Pablo Marquez-Neila, Mathieu Salzmann, and Pascal Fua.
\newblock Learning to fuse 2{D} and 3{D} image cues for monocular body pose
  estimation.
\newblock In {\em International Conference on Computer Vision}, 2017.

\bibitem{Tome_2017_CVPR}
Denis Tome, Chris Russell, and Lourdes Agapito.
\newblock Lifting from the deep: Convolutional 3{D} pose estimation from a
  single image.
\newblock In {\em IEEE Conference on Computer Vision and Pattern Recognition},
  2017.

\bibitem{tung2017}
Hsiao-Yu~Fish Tung, Adam~W Harley, William Seto, and Katerina Fragkiadaki.
\newblock Adversarial inverse graphics networks: {L}earning 2d-to-3d lifting
  and image-to-image translation from unpaired supervision.
\newblock In {\em International Conference on Computer Vision}, 2017.

\bibitem{tsne}
Laurens van~der Maaten and Geoffrey Hinton.
\newblock Visualizing data using t-sne.
\newblock In {\em Journal of Machine Learning Research}.

\bibitem{wei2016}
Shih-En Wei, Varun Ramakrishna, Takeo Kanade, and Yaser Sheikh.
\newblock Convolutional pose machines.
\newblock In {\em IEEE Conference on Computer Vision and Pattern Recognition},
  2016.

\bibitem{3dinterpreter}
Jiajun Wu, Tianfan Xue, Joseph~J Lim, Yuandong Tian, Joshua~B Tenenbaum,
  Antonio Torralba, and William~T Freeman.
\newblock Single image 3d interpreter network.
\newblock In {\em European Conference on Computer Vision (ECCV)}, 2016.

\bibitem{Nie_2017_ICCV}
Bruce Xiaohan~Nie, Ping Wei, and Song-Chun Zhu.
\newblock Monocular 3d human pose estimation by predicting depth on joints.
\newblock In {\em International Conference on Computer Vision}, 2017.

\bibitem{zhou2016sparseness}
Xiaowei Zhou, Menglong Zhu, Kosta Derpanis, and Kostas Daniilidis.
\newblock Sparseness meets deepness: 3{D} human pose estimation from monocular
  video.
\newblock In {\em IEEE Conference on Computer Vision and Pattern Recognition},
  2016.

\end{thebibliography}
}

\end{document}